# OPTIMIZE FLIP ANGLE SCHEDULES IN MR FINGERPRINTING USING REINFORCEMENT LEARNING

*Shenjun Zhong*[1,2], *Zhifeng Chen*[3], *Zhaolin Chen*[1,4]

[1]Monash Biomedical Imaging, Monash University, Australia
[2]National Imaging Facility, Australia
[3]Neusoft Medical Systems, Hangzhou, China
[4]Faculty of Information Technology, Monash University, Australia

**ABSTRACT**

*Magnetic Resonance Fingerprinting (MRF) leverages transient-state signal dynamics generated by the tunable acquisition parameters, making the design of an optimal, robust sequence a complex, high-dimensional sequential decision problem, such as optimizing one of the key parameters, flip angle. Reinforcement learning (RL) offers a promising approach to automate parameter selection, to optimize pulse sequences that maximize the distinguishability of fingerprints across the parameter space. In this work, we introduce an RL framework for optimizing the flip-angle schedule in MRF and demonstrate a learned schedule exhibiting non-periodic patterns that enhances fingerprint separability. Additionally, an interesting observation is that the RL-optimized schedule may enable a reduction in the number of repetition time, potentially accelerate MRF acquisitions.*

## 1. INTRODUCTION

The acquisition phase of Magnetic Resonance Fingerprinting(MRF), characterized by the pseudo-random variation of parameters, presents a sophisticated optimal control problem. Some works use empirical sequence design for MRF acquisition, like QALAS that uses an interleaved Look-locker sequence with T2 preparation pulse. There has been exploration of optimizing sequence design for MRF, mostly using the Cramér-Rao lower bound (CRLB) optimization [1]. Zhao et al. provides a general framework for flip angle (FA) and Repetition Time (TR) optimization in MRF [2]; Asslander et al. proposed an optimal-control style framework to design FA and TR patterns to minimize noise via CRLB [3]; Lee et al. applied an automatic differentiation on bloch-simulations to compute gradients of CRLB with respect to FA schedule and TR [4]. More recently, Slioussarenko et al. Optimized the FA and TE parameters of a FLASH-based MRF sequence and achieved 30% scan duration reduction [5].

Reinforcement learning (RL) is uniquely positioned to address this by automating the selection of parameters [6], e.g. flip angle to generate an optimized pulse sequence that maximizes the distinctness of fingerprints across the parameter space. In this work, we propose a reinforcement learning framework that interacts with a GPU-accelerated Extended Phase Graphs [7] simulator to learn the FA schedule.

The contributions of this works are: (1) We propose, for the first time, a reinforcement learning framework to optimize the flip-angle schedule in MR fingerprinting; (2) We demonstrate a learned flip-angle schedule that exhibits non-periodic patterns and improves fingerprint separability; (3) Our results indicate that the RL-optimized schedule may allow a reduction in the number of TRs, potentially accelerating MRF acquisition.

## 2. METHOD

### 2.1. Fast EPG Simulator

We implement a multi-level parallelism strategy, enabling intra-simulation parallelism (parallelising the execution of one EPG simulation) and inter-simulation parallelism (running multiple simulations simultaneously). Within this framework, we implemented a Steady-State Free Precession (SSFP) sequence [8] using the accelerated EPG implementation for validation purposes.

To enable integration with reinforcement learning, the EPG simulation must be interactive. In this work, the accelerated EPG simulator is encapsulated as a Gym environment, providing a standardized interface for interaction with the training process.

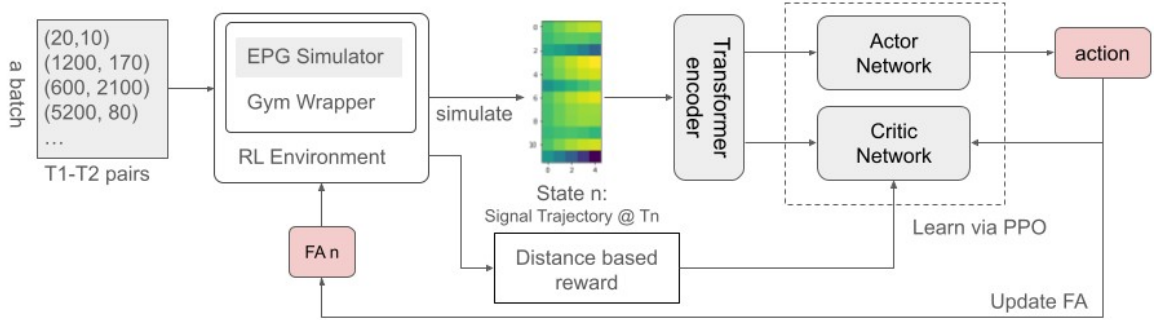

Figure 1: Overview of Workflow and Architecture Design.

## 2.2. Optimizing Flip Angles via PPO

**Environment Setup**: The interactive environment defines a Gymnasium environment for RL-based flip angle optimization in MR fingerprinting. The agent adjusts flip angles over multiple TRs, and the environment simulates parallel EPG sequences with varied T1/T2 parameters sampled from a dictionary range. Observations are echo density matrices and rewards come from margin-based functions that encourage feature diversity/dissimilarity. The actions correspond to small adjustments to the previous flip angle, in this case, only a single-degree adjustment is allowed per TR.

**Policy Network Architecture**: The policy network uses a Transformer encoder as the feature extractor, processing time-series echo signals. The encoder includes configurable parameters (embedding dimension, number of attention heads, and layers). The extracted features are passed through a multi-layer perceptron (MLP) that outputs separate policy and value heads for the actor-critic architecture.

**Reward Function**: The reward function is design to maximize the dissimilarities between the MRF signal trajectory. We introduce a margin-based reward function to maximize pairwise dissimilarity across all time series in a batch. The reward combines a positive term for pairs exceeding a margin threshold and a negative term for pairs below it, normalized by the number of pairs.

Given a batch of time series features $F = f_1, f_2, ..., f_N$, we compute a pairwise dissimilarity matrix $D \in R^{N \times N}$ using normalized dot product for measuring similarity. The reward is:

$$R = \frac{1}{|P|} \left( \sum_{(i,j) \in P} max(0, D_{ij} - \tau) - max(0, \tau - D_{ij}) \right)$$

where $\tau$ is the margin threshold, P is the set of all unique pairs, excluding self-comparisons, and $|P| = N(N-1)/2$.

The first term encourages pairs to exceed the margin (dissimilarity), while the second penalizes pairs below the margin (similarity). Normalization by $|P|$ stabilizes the reward across batch sizes. This formulation promotes diversity across all pairs, unlike approaches that only reward pairs above the threshold, ensuring the learned sequences are maximally dissimilar.

**PPO Optimization**: Training uses Proximal Policy Optimization (PPO) with a Transformer-based actor-critic policy [9]. PPO [10] is an on-policy algorithm that updates the policy using collected trajectories. The agent collects n timesteps of experience by interacting with the EPG simulation environment. At each step, the agent observes the current echo signal state, selects an action (flip angle adjustment) from a discrete action space, and receives a reward based on the dissimilarity between time series in the batch.

**Training**: for training, each batch is randomly sampled from 82x82 T1-T2 pairs, where the dissimilarity is measured in a batch. The FA schedule at the end of the training or the best checkpoint is selected for the evaluation in this work.

## 3. RESULTS

In our experiments, we evaluate the dissimilarity between the generated signal trajectories and those in the MRF dictionary across a broad T1–T2 parameter space. PPO is trained using the following hyperparameters: learning rate 3e10-4, rollout length 2048, batch size 64, 10 optimization epochs, discount factor 0.99, GAE λ = 0.95, clip range 0.2, entropy coefficient 0.01, value loss coefficient 0.5, and maximum gradient norm 0.5. The policy network employs a transformer-based feature extractor (128-dimensional embedding, 8 heads, 2 layers) followed by an MLP head with a 128-dimensional hidden layer. The model was trained on Nvidia A100 GPU, for 500,000 iterations, and the best performed checkpoint (around 360,000 iterations) is selected.

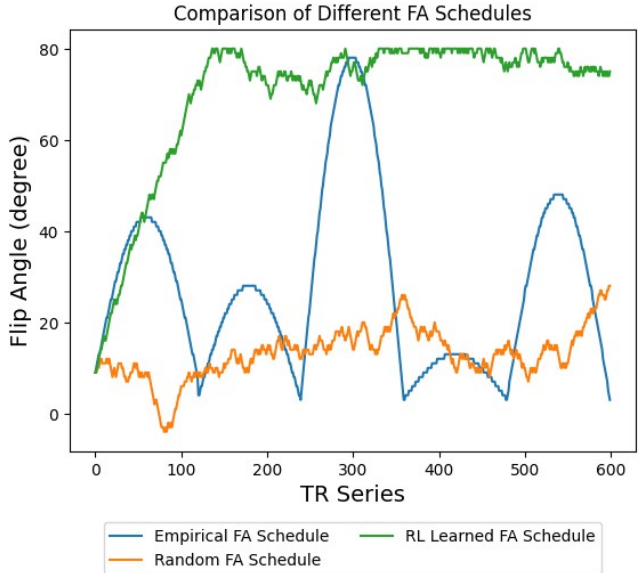

Figure 2: Different Flip Angle Schedules, Including Empirical Design, Random FA and FA Learned via RL Training.

**RL learned FA schedule.** Figure 2 shows the flip-angle schedule learned through reinforcement learning, compared with a randomly generated schedule and an empirically designed one. Whereas the empirical schedule is constructed from combinations of sinusoidal patterns, the learned schedule does not exhibit periodic behavior. Instead, it gradually increases the flip angle from an initial value of 9° toward the upper limit of 80°, with additional fluctuations superimposed.

**MRF Signal Dissimilarity Comparison.** The goal of the generated MRF signal trajectories is to maximize dissimilarity across tissue parameters, improving dictionary matching accuracy. Figure 3 compares average pairwise distances among fingerprints across a wide range of T1–T2 values. Fingerprints generated with the RL-learned FA schedule show the highest distances, suggesting it may outperform the empirical design for 100–300 TRs. Figure 4 further shows that the dissimilarity distributions differ, with the learned FA schedule yielding slightly higher separations in some regions.

**Visualization of the Fingerprints**. We sampled 128 T1–T2 combinations on a grid and visualized the corresponding fingerprints in the generated dictionary. Figures 5(a) and 5(b) show fingerprints with varying T1 values while T2 is fixed to a short value. The learned schedule produces fingerprints with improved separability, even when using shorter TR counts (around 200–300). This suggests that some TRs may be redundant, indicating potential for reducing the total number of TRs in MRF acquisitions. A similar trend is observed in Figures 5(c) and 5(d), where T1 is fixed and T2 varies.

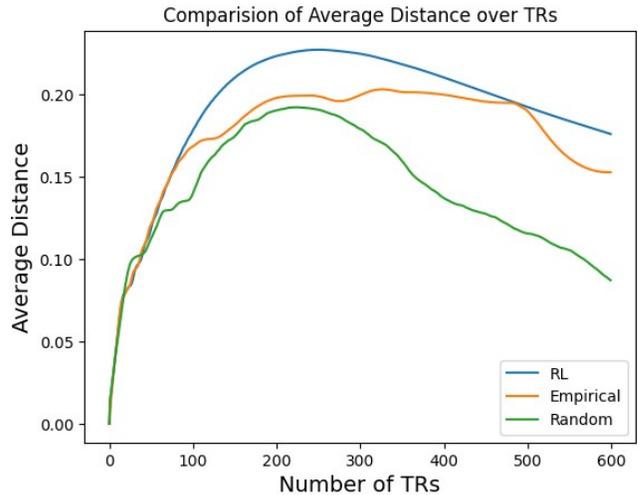

Figure 3: Comparing the Dissimilarities over TRs between Random FA, empirical FA and RL Learnt FA Schedules.

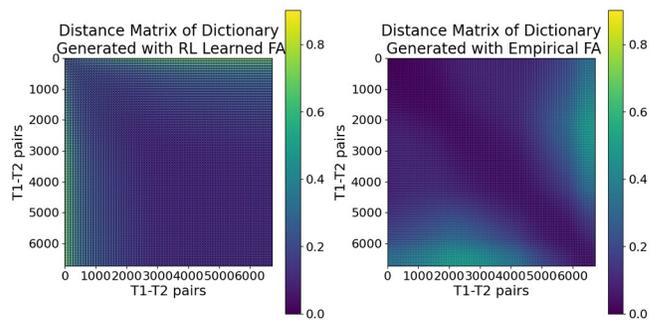

Figure 4: Comparison of Distance Matrix of the Fingerprint Dictionary Using the Emperical and RL Learned FAs.

## 5. CONCLUSION

In this work, we proposed a reinforcement learning framework for optimizing the flip-angle schedule in MR fingerprinting. The learned, non-periodic flip-angle pattern improves fingerprint separability in our simulations. Moreover, the results indicate substantial redundancy in the number of TRs when using the RL-optimized schedule, suggesting opportunities to reduce TR count and accelerate MRF acquisitions. Besides optimizing the flip-angle schedule, the same framework can be applied to optimize other MRF acquisition parameters. However, acquisition artifacts and noise were not modeled in this study; these factors will be addressed in future work.

applied to the fast 3D quantification of fat fraction and water T1 in the thigh muscles. Magnetic Resonance in Medicine, 93(6),

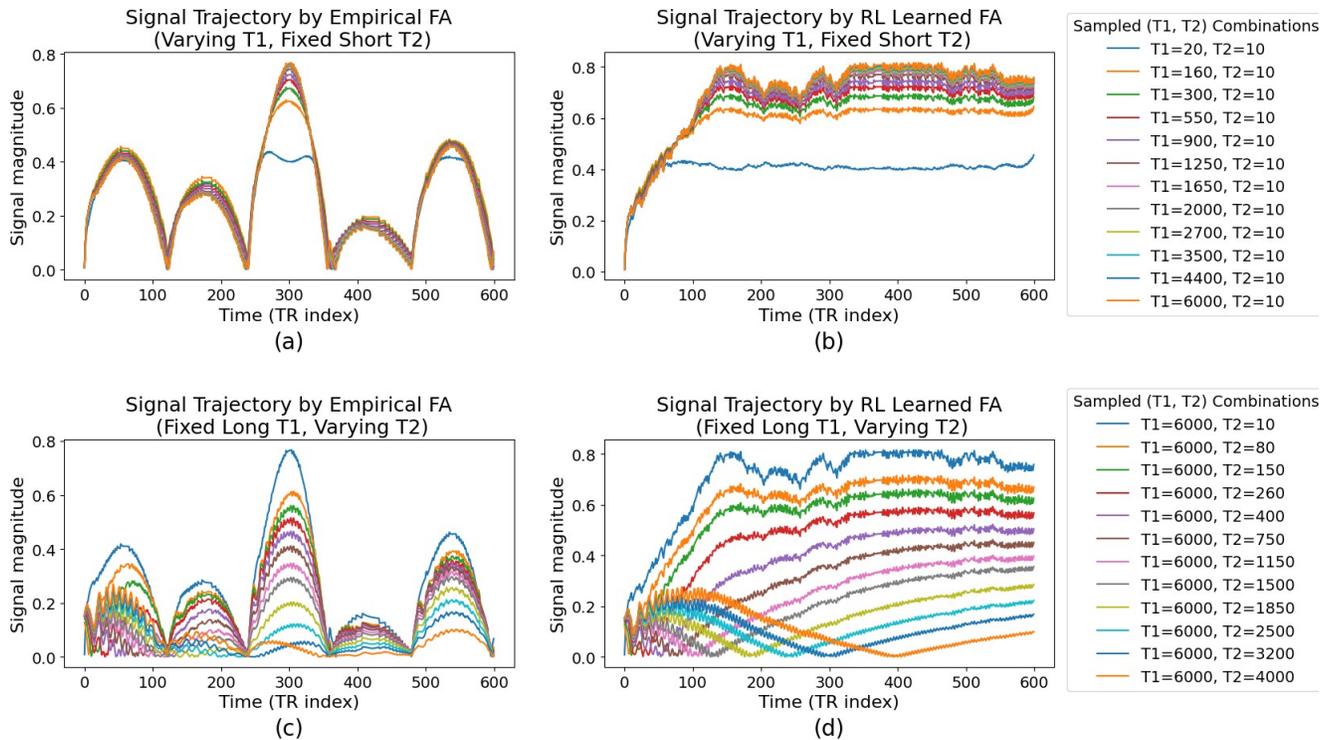

Figure 5: Visualization And Comparison of Sampled Fingerprints with Varying T1 and T2.